# Deriving Hematological Disease Classes Using Fuzzy Logic and Expert Knowledge: A Comprehensive Machine Learning Approach with CBC Parameters

Salem Ameen
School of Science, Engineering and Environment
*University of Salford*
Manchester, United Kindom
s.a.ameen1@salford.ac.uk

Ravivarman Balachandran
School of Science, Engineering and Environment
*University of Salford*
Manchester, United Kindom
r.balachandran@edu.salford.ac.uk

Theodoros Theodoridis
School of Science, Engineering and Environment
*University of Salford*
Manchester, United Kindom
t.theodoridis@salford.ac.uk

*Abstract*

In the intricate field of medical diagnostics, capturing the subtle manifestations of diseases remains a challenge. Traditional methods, often binary in nature, may not encapsulate the nuanced variances that exist in real-world clinical scenarios. This paper introduces a novel approach by leveraging Fuzzy Logic Rules to derive disease classes based on expert domain knowledge from a medical practitioner. By recognizing that diseases don't always fit into neat categories, and that expert knowledge can guide the fuzzification of these boundaries, our methodology offers a more sophisticated and nuanced diagnostic tool.

Using a dataset procured from a prominent hospital, containing detailed patient blood count records, we harness Fuzzy Logic Rules—a computational technique celebrated for its ability to handle ambiguity. This approach, moving through stages of fuzzification, rule application, inference, and ultimately defuzzification, produces refined diagnostic predictions. When combined with the Random Forest classifier, the system adeptly predicts hematological conditions using Complete Blood Count (CBC) parameters.

Preliminary results showcase high accuracy levels, underscoring the advantages of integrating fuzzy logic into the diagnostic process. When juxtaposed with traditional diagnostic techniques, it becomes evident that Fuzzy Logic, especially when guided by medical expertise, offers significant advancements in the realm of hematological diagnostics. This paper not only paves the path for enhanced patient care but also beckons a deeper dive into the potentialities of fuzzy logic in various medical diagnostic applications.

**Keywords**, Hematological Diagnostics, Machine Learning, Fuzzy Logic, Diagnostic Accuracy and Bias Mitigation.

## 1. Introduction

In the rapidly advancing field of digital medicine, the synergy between computational methods and clinical diagnostics is becoming increasingly paramount. As we journey deeper into this era, we confront a pressing conundrum: while traditional diagnostic techniques are adept at categorizing diseases in a binary fashion, they often falter in capturing the subtle, intricate gradations of disease manifestations. This shortcoming prompts an essential inquiry: How might we encapsulate the multifarious shades of medical conditions, ensuring that no nuanced symptom remains obscured?

Enter Fuzzy Logic, a computational paradigm pioneered by Lotfi Zadeh in 1965 [1]. Renowned for its ability to handle ambiguity, Fuzzy Logic transcends binary true-or-false classifications, making it particularly well-suited for the medical field where symptoms and conditions exist on a vast and interconnected spectrum. Fuzzy Logic shines a light on the gray areas of diagnosis, offering a more nuanced lens through which to view patient data. However, its true potential is realized when combined with the expert knowledge of medical practitioners. By incorporating domain-specific insights from doctors, we can refine the fuzzy boundaries between disease categories, accurately identify healthy individuals, and recognize when test results are ambiguous and may require retesting.

This paper sets out to leverage augmented Fuzzy Logic for the analysis of a range of hematological conditions, from anemia to leukemia and beyond. It also places a strong emphasis on accurately categorizing healthy individuals and identifying cases where the diagnostic results are inconclusive. These latter cases, termed "No Disease Detection," highlight situations where the blood parameters do not provide clear evidence of disease, yet the results are not confidently normal either, signaling the potential need for further testing. Drawing on a meticulously curated dataset from a renowned Sri Lankan Hospital, our research develops a Fuzzy Logic System enriched with medical expertise. The aim is to predict the onset and progression of hematological conditions with unprecedented precision, accurately identify healthy states, and flag cases that may require additional investigation.

As we progress through this paper, we will unveil our research journey—detailing our methodologies, exploring the intricacies of our dataset, presenting our results, and comparing our approach to traditional diagnostic techniques. Special attention will be given to how our system differentiates between various hematological conditions, healthy states, and "No Disease Detection" cases, shedding light on this vital aspect of medical diagnostics. By intertwining the depth of medical knowledge with the capabilities of Fuzzy Logic and machine learning, we endeavor to introduce a new era of diagnostic tools that truly capture the complexity of medical data, the certainty of health, and the ambiguity that sometimes requires a second look.

## 2. Literature Review

The healthcare sector's burgeoning relationship with machine learning (ML) has ushered in an era of enhanced diagnostic capabilities, especially concerning hematological disorders [2-4]. machine learning, an integral branch of artificial intelligence (AI), facilitates automated learning from patterns, allowing systems to formulate decisions without explicit programming [1].

Several studies have underscored the promise of machine learning in analyzing complete blood counts (CBC) for disease diagnosis. Researchers suggested that machine learning techniques eclipse traditional methods in blood cell estimation, forecasting, and categorization [5, 6]. Building on this, Buttarello & Plebani (2008) emphasized the routine nature of CBCs in medical tests and the potential for automated systems to streamline the counting process [7]. The interplay between fuzzy logic and machine learning has attracted considerable attention for its potential in processing imprecise data and enhancing prediction accuracy [8-10].

Gholamzadeh et al. (2018) discussed enhancing predictive accuracy in medical diagnostics through the role of fuzzy logic, underscoring the capacity of fuzzy systems to deal with data that isn't strictly binary [8]. The uniqueness of fuzzy logic comes to the fore when deriving classes based on domain knowledge from doctors, emphasizing its crucial role in medical diagnostics [6]. In a similar vein, Susanto et al. (2021) employed a fuzzy-centric approach for predicting thalassemia diseases, illustrating the heightened diagnostic results when integrating fuzzy systems with machine learning methodologies [11].

Researchers have explored the integration of various machine learning algorithms to optimize decision-making in medical diagnostics [12].

In the domain of hematological diagnostics, significant strides have been made in segmenting white blood cells (WBCs) using advanced algorithms such as watershed, k-means, and fuzzy c-means [13].

These techniques form the bedrock for early disease detection. M. Burgermaster et al, (2020) recognized the mutual benefits arising from the synergy of clinical experts and machine learning models in healthcare, shedding light on the importance of expert domain knowledge [14]. However, the integration of machine learning in healthcare presents challenges. Johnson et al. (2022) highlighted issues like data non-stationarity, model interpretability, and the quest for appropriate data representations [15, 16].

Balancing these challenges requires a collaborative ethos between clinicians and machine learning researchers for a truly synergistic impact [17, 18].

In conclusion, the intersection of machine learning with healthcare diagnostics, notably in hematological assessments, is poised to redefine the landscape.

# 3. Dataset Description

## 3.1 Collection and Usage Rights:
The dataset was sourced from a reputed healthcare institution, adhering to stringent privacy standards. All data has been anonymized and sanitized, ensuring that no personally identifiable information of patients is present.

## 3.2 Generation of Dataset:
Contemporary medical institutions deploy an array of specialized instruments for blood analysis in various departments like Hematology, Biochemistry, Microbiology, Immunology, and Histopathology. For this study, data was singularly derived from the Hematology department. The primary analytical tool employed was the Full Blood Count (FBC), also known as the Complete Blood Count (CBC). This instrument efficiently generates .CSV (comma-separated values) files, which were subsequently used to formulate our dataset.

## 3.3 Key Parameters of the Dataset:
- Sample ID: A unique identifier assigned to each blood count analysis, ensuring traceability to specific medical conditions.
- WBC (White Blood Cell): A pivotal metric, WBCs are instrumental in defending the body against infections. They include five significant components:
- Neu (Neutrophils): Act as the first line of defense against bacterial and fungal infections.
- Lym (Lymphocytes): Cells responsible for producing antibodies, enhancing the body's resilience against pathogens.
- Mon (Monocytes): Primarily involved in breaking down bacteria.
- Eos (Eosinophils): Primarily target parasites, cancer cells, and oversee allergic reactions.
- Bas (Basophils): Indicative of potential infections and play a role in immune responses.
- RBC (Red Blood Cell): Responsible for oxygen transport within the body and aids in expelling carbon dioxide back to the lungs.
- HGB (Hemoglobin): An iron-laden protein crucial for transporting oxygen in the bloodstream.
- HCT (Hematocrit): Denotes the percentage of blood volume occupied by red blood cells. Variations can indicate potential medical anomalies.
- PLT (Platelets): Essential for blood clotting and the healing of injuries.
- Gender & Ref. Group: While gender contributes to certain classification tasks, the 'Ref. Group' divides data into specific categories like adult male, adult female, child, and neonate.
- Age: Enhances predictive accuracy, especially when coupled with gender.
- Cellular Messages (WBC, RBC, PLT): Advanced analyzers furnish intricate details on cell types, capturing nuances in cell sizes, patterns, and potential indicators of diseases. Integrating this data into the machine learning framework can enhance the precision of disease diagnostic models.

## 4. Methods

### 4.1 Fuzzy Logic Implementation

**Overview**: Our primary step was the implementation of fuzzy logic, influenced by the invaluable domain knowledge provided by medical professionals. This approach aimed to capture the inherent uncertainties in medical diagnoses, making our model's predictions more nuanced and patient-centric.

**Fuzzification**: Leveraging the skfuzzy library, we transformed conventional values into fuzzy equivalents through membership functions. These functions determined to what extent an element was part of a set. We explored Gaussian, triangular, and trapezoidal memberships to determine which resonated best with our dataset.

**Rule Base**: Our robust set of IF-THEN rules, constructed with the aid of expert knowledge, defined the intricate relationship between our inputs and outputs.

**Inference System**: We employed the Mamdani method to amalgamate our rules, resulting in a unified output function.

**Defuzzification**: To transition from fuzzy logic to conventional data analysis, we used the centroid method to extract clear-cut values from the fuzzy outputs.

### 4.2 Data Preprocessing for Machine Learning

**Data Cleaning**: Post the fuzzy logic phase, ensuring the integrity of our medical data became paramount. We removed any entries that were missing, incomplete, or redundant. Outliers were identified using the Interquartile Range (IQR) method and were managed accordingly.

**Feature Selection**: Relying on domain expert advice, we focused on clinically significant features such as 'WBC', 'HGB', 'HCT', 'PLT', and 'Age'.

**Normalization**: All numerical values underwent normalization via the Min-Max scaling technique, ensuring a consistent [0,1] range.

**Dataset Resampling**: The distribution of samples among various disease categories in our initial training dataset is summarized in Table 1

| Disease/Class Category | No. of Samples |
| --- | --- |
| **Iron Deficiency Anemia - IDA / Sickle Cell Anemia / Acute Blood Loss** | 850 |
| **Healthy** | 804 |
| **No Disease Detected** | 721 |
| **Septicemia / Urine Tract Infections - UTI** | 261 |
| **Septicemia** | 63 |
| **Other Viral Fevers / Idiopathic Thrombocytopenic Purpura - ITP** | 34 |
| **Other Viral Fevers** | 33 |
| **Dengue** | 28 |
| **Chronic Kidney Disease - CKD** | 14 |
| **Pancytopenia** | 8 |
| **Polycythemia** | 1 |

Table 1: Distribution of Samples Among Disease Categories in the Initial Training Dataset

Due to the imbalances in our training data, we employed the Random Over Sampler technique. This resulted in each disease category having 850 samples, ensuring a balanced dataset for model training.

### 4.3 Machine Learning Model

**Choice of Model**: We utilized the Random Forest Classifier from the scikit-learn library.

**Rationale**: The Random Forest model was strategically selected for its capability to handle vast datasets, its resilience to overfitting, and its ability to provide feature importance insights.

**Training and Validation**: The dataset was divided in a 70-30 ratio. The training set refined the model, while the validation set evaluated its performance.

**Performance Metrics**: We critically assessed the model using metrics like accuracy, precision, recall and F1 score to ensure a comprehensive evaluation of its capabilities.

## 5. Results

### 5.1 Model Configuration

The Random Forest Classifier was optimized with the following hyperparameters: Criterion set to 'Gini', Max Depth at 6, and Number of Estimators at 100. These were determined after a comprehensive grid search to achieve the best balance between model complexity and performance.

### 5.2 Performance on Training Dataset

The model, trained on 9,350 samples, showcased:

- Near-perfect precision, recall, and F1-score for 'IDA or SCA or ABL' and 'Healthy'.

- A weighted average accuracy of 99%, highlighting its potential generalizability.

- The heatmap of the performance metrics for the training dataset, depicted in Figure 1, offers an in-depth look into the model's predictions across the various disease categories.

| Categories | Precision | Recall | F1-score |
|---|---|---|---|
| IDA or SCA or ABL | 1.00 | 1.00 | 1.00 |
| Healthy | 1.00 | 1.00 | 1.00 |
| No Disease | 0.96 | 1.00 | 0.98 |
| Sep or UTI | 0.99 | 1.00 | 1.00 |
| CLCD or ABL | 1.00 | 0.91 | 0.95 |
| OVF or ITP | 1.00 | 1.00 | 1.00 |
| OVF | 0.99 | 1.00 | 0.99 |
| Pan | 1.00 | 1.00 | 1.00 |
| CKD | 1.00 | 1.00 | 1.00 |
| Sep | 1.00 | 1.00 | 1.00 |
| Pol | 0.98 | 1.00 | 0.99 |

Figure 1: Performance on Training Dataset

### 5.3 Performance on Testing Dataset

On the independent testing dataset of 940 samples, the model achieved an overall accuracy of 97%. Commendable precision and recall were noted for most categories. However, 'CLCD or ABL' experienced some misclassifications, evident from its recall of 89%.

The model demonstrated exceptional capability in accurately identifying 'Healthy' cases, achieving 100% precision and recall, ensuring that healthy individuals are confidently identified. In the 'No Disease Detection' category, the model achieved a precision of 95% and a recall of 100%, highlighting its proficiency in identifying cases that may require further investigation.

The confusion matrix for this dataset provided insights into the model's predictions for each category, revealing areas of high performance and aspects that need further attention. These results and insights are presented in Figure 2.

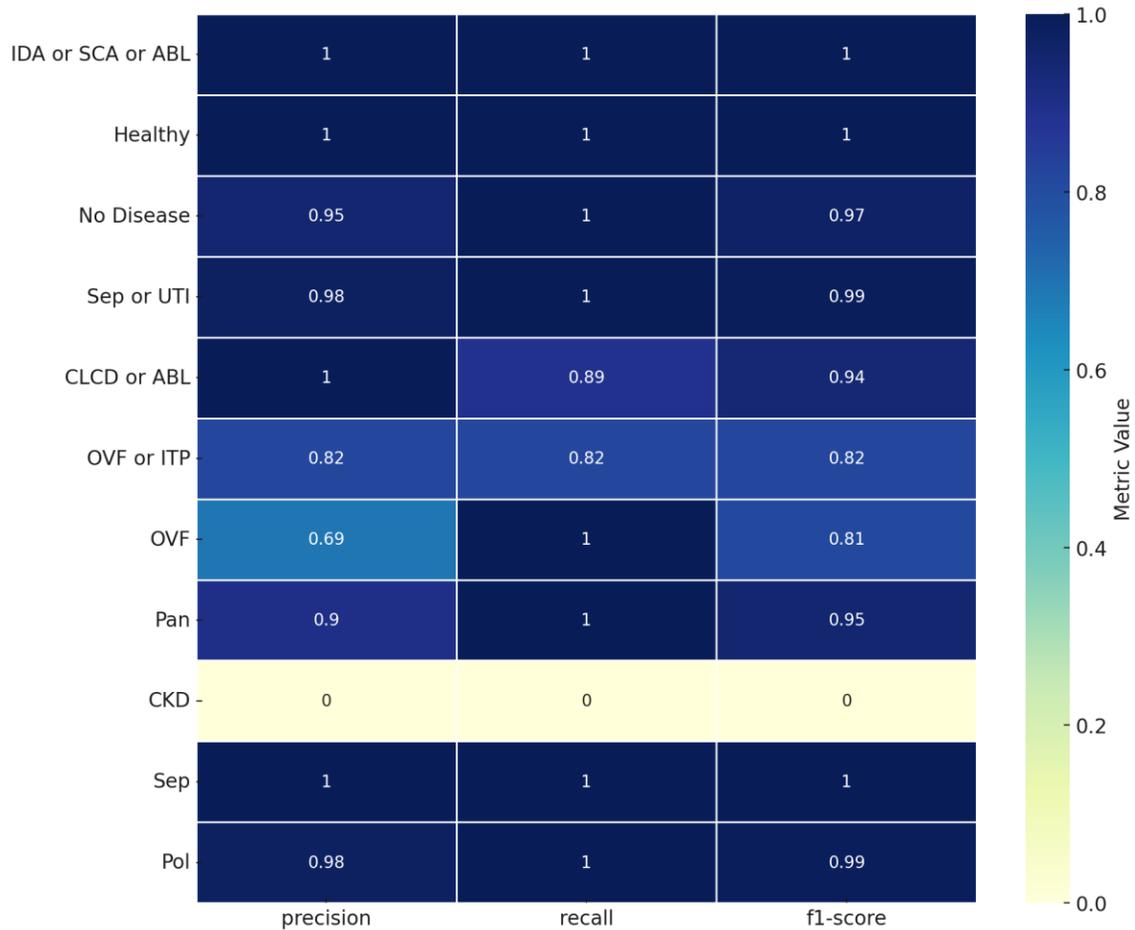

Figure 2: Performance on Testing Dataset

## 6. Discussion and Analysis

The Random Forest Classifier showcased a promising performance, particularly in identifying 'Healthy' cases and instances requiring further investigation. The near-perfect scores in these categories emphasize the model's potential utility in a clinical setting, providing reliable and crucial information to healthcare professionals.

The model's performance in the 'IDA or SCA or ABL' and 'Healthy' categories was noteworthy, highlighting its applicability for early disease diagnosis. However, the 'CLCD or ABL' category presented challenges, especially in differentiating this disease from others, which can be attributed to the inherent similarities in hematological markers. This points to an area where the model could potentially be improved.

The dataset's imbalance, particularly in the 'CKD' category, poses a significant limitation, emphasizing the necessity for a balanced dataset to achieve optimal classification results. Despite this, the minimal performance disparity between the training and testing datasets illustrates the model's robustness and its ability to generalize well across different data samples.

Figure 3, depicting the heatmap of the confusion matrix for the testing dataset, provides additional insights into the model's performance across various disease categories. This visual representation is crucial for understanding the model's behavior, pinpointing areas of high performance, and identifying potential avenues for model enhancement.

| True \ Predicted | IDA or SCA or ABL | Healthy | No Disease | Sep or UTI | CLCD or ABL | OVF or ITP | OVF | Pan | CKD | Sep | Pol |
|---|---|---|---|---|---|---|---|---|---|---|---|
| IDA or SCA or ABL | 5 | 0 | 0 | 0 | 0 | 0 | 0 | 0 | 0 | 0 | 0 |
| Healthy | 0 | 21 | 0 | 0 | 0 | 0 | 0 | 0 | 0 | 0 | 0 |
| No Disease | 0 | 0 | 268 | 0 | 0 | 0 | 0 | 0 | 0 | 0 | 0 |
| Sep or UTI | 0 | 0 | 0 | 284 | 0 | 0 | 0 | 0 | 0 | 0 | 0 |
| CLCD or ABL | 0 | 0 | 14 | 5 | 213 | 2 | 4 | 0 | 0 | 0 | 2 |
| OVF or ITP | 0 | 0 | 0 | 0 | 0 | 9 | 1 | 1 | 0 | 0 | 0 |
| OVF | 0 | 0 | 0 | 0 | 0 | 0 | 11 | 0 | 0 | 0 | 0 |
| Pan | 0 | 0 | 0 | 0 | 0 | 0 | 0 | 9 | 0 | 0 | 0 |
| CKD | 0 | 0 | 0 | 0 | 1 | 0 | 0 | 0 | 0 | 0 | 0 |
| Sep | 0 | 0 | 0 | 0 | 0 | 0 | 0 | 0 | 0 | 3 | 0 |
| Pol | 0 | 0 | 0 | 0 | 0 | 0 | 0 | 0 | 0 | 0 | 87 |

Figure 3: Confusion Matrix

In conclusion, while the model displays significant potential in the realm of medical diagnostics, it is vital to acknowledge its existing limitations. Prioritizing areas for refinement and addressing the highlighted challenges will be crucial in enhancing its performance and reliability in real-world applications.

## 7. Conclusion

The field of medical diagnostics is constantly evolving, and the integration of advanced methodologies such as machine learning is shaping its future. This study has uniquely combined the principles of fuzzy logic with machine learning, resulting in a more refined categorization of complex hematological data. The inclusion of fuzzy logic significantly enhances the quality of input fed into our Random Forest classifier, which is reflected in the results.

With an impressive accuracy rate of 97% on the testing dataset, this study goes beyond showcasing numerical success; it highlights the potential for early and precise disease detection. Such capabilities are crucial for initiating timely medical interventions. The model's remarkable performance, particularly in identifying diseases with overlapping clinical markers, affirms the enhanced robustness achieved through fuzzy logic preprocessing.

While we acknowledge the successes of this study, it is important to recognize the challenges encountered along the way. Addressing issues such as dataset imbalances and its regional focus provides us with a broader perspective and identifies areas for improvement.

In essence, this research not only highlights the importance of machine learning in medical diagnostics but also underscores the value of domain-specific expertise, realized through the integration of fuzzy logic, in improving predictive accuracy.

## 8. Future Work

Hematological diagnostics is a vast domain, brimming with opportunities for deeper exploration and innovation:

**Refining Fuzzy Logic**: As medical knowledge continues to evolve, there is potential to enhance and update the fuzzy logic rules to ensure they remain aligned with the latest domain insights.

**Predicting Specific Blood Cell Counts**: Moving beyond disease diagnosis, future research could focus on predicting specific blood cell counts, providing crucial information for early medical intervention.

**Addressing Class Imbalances**: Although our current model has demonstrated promising results, there is room for improvement, especially in handling imbalanced classes. Employing cost-sensitive learning techniques, which assign different misclassification costs to different classes, can help the model better understand and represent underrepresented classes. This is particularly important in medical diagnostics, where overlooking a rare but severe condition can have severe implications. Investigating other dataset augmentation methods to enhance the representation of minority classes is also a viable path forward.

**Broadening Data Integration**: Incorporating additional data types, such as patient histories, can provide a more comprehensive context for predictions.

**Expanding Model Horizons**: While the Random Forest classifier has performed well, exploring an ensemble of models could offer a broader and more holistic diagnostic perspective.

 **Ethical Modeling and Bias Mitigation**: As the field progresses, maintaining a focus on ethical modeling and proactive bias mitigation remains paramount.

**Paving the Path for Global Application:** Future developments should aim to adapt the model to suit the diverse hematological profiles of global populations, ensuring its applicability and relevance worldwide.

As we look forward, the intertwined challenges and prospects highlight a bright future for hematological diagnostics, promising advancements and innovations in the years to come.

## Author Contributions

**Salem Ameen**: Engaged in software development, data analysis, manuscript writing, and validation processes.

**Ravivarman Balachandran**: Facilitated data acquisition, applied his extensive experience working in hospitals to define the fuzzy logic outputs, contributed to the software development, and conducted investigations.

**Theo Theodoridis**: Participated in the validation processes, provided necessary resources, reviewed and edited the manuscript, and administered project tasks.

## Acknowledgments

We extend our sincere gratitude to the hospital for providing access to the dataset and to the medical professionals who shared their valuable experience, significantly contributing to the decision-making process reflected in this work.

# Appendix

## A. Fuzzy Logic Rules for Disease Detection

The following are the fuzzy logic rules that were used to classify the diseases based on haematological data:

- Septicaemia/Urine Tract Infections (UTI): Detected when WBC is high, HGB is normal, HCT is normal, and PLT is normal.
- Septicaemia: Detected when WBC is high, HGB is normal, HCT is normal, and PLT is high.
- Dengue: Detected when WBC is either low or normal, HGB is high, HCT is high, and PLT is low.
- Other Viral Fevers: Detected under two conditions:
- WBC is low, HGB is normal, HCT is normal, and PLT is low.
- WBC is normal, HGB is normal, HCT is normal, and PLT is low.
- Iron Deficiency Anaemia - IDA/Sickle Cell Anaemia/Acute Blood Loss: Detected when WBC is normal, HGB is low, HCT is low, and PLT is normal.
- Polycythaemia: Detected when WBC is normal, HGB is high, HCT is high, and PLT is normal.
- Pancytopenia: Detected when WBC is low, HGB is low, HCT is low, and PLT is low.
- Chronic Kidney Disease - CKD: Detected when WBC is high, HGB is low, HCT is low, and PLT is low.
- Chronic Liver Cell Disease - CLCD/Acute Blood Loss: Detected when WBC is normal, HGB is low, HCT is low, and PLT is low.
- Healthy: Detected when all values, WBC, HGB, HCT, and PLT, are normal.

## B. Fuzzy Sets Definition

For the purpose of this study, fuzzy sets were defined for each haematological parameter:

- WBC (White Blood Cells):
    - Low: 0 to 4
    - Normal: 4 to 10
    - High: 10 to 60
- HGB (Haemoglobin):
    - Low: 0 to 12
    - Normal: 12 to 16
    - High: 16 to 50
- HCT (Haematocrit):
    - Low: 0 to 35
    - Normal: 35 to 54
    - High: 54 to 100
- PLT (Platelet Count):
    - Low: 0 to 100

- Normal: 100 to 450
- High: 450 to 1500

The above definitions of fuzzy sets were utilized to transform the raw haematological data into a format suitable for fuzzy rule processing.

### C. Sample Disease Checks

A random sample were tested using the fuzzy logic system. The result is presented below:

For Sample 0 from the dataset:

- WBC Value: ms_df['WBC'][0]
- HGB Value: ms_df['HGB'][0]
- HCT Value: ms_df['HCT'][0]
- PLT Value: ms_df['PLT'][0]

Predicted Disease: [CDK]

### D. Code Implementation

The fuzzy logic rules and the machine learning model were implemented using Python. For a detailed code implementation, readers can access the link:

https://github.com/SalemAmeen/DiseasePredictionModal/blob/main/Disease%20Prediction%20Modal.ipynb